# Igea: a Decoder-Only Language Model for Biomedical Text Generation in Italian

Tommaso Mario Buonocore[1], Simone Rancati[1], and Enea Parimbelli[1]

Dept. of Electrical, Computer and Biomedical Engineering, University of Pavia, Pavia, Italy

**Abstract**

The development of domain-specific language models has significantly advanced natural language processing applications in various specialized fields, particularly in biomedicine. However, the focus has largely been on English-language models, leaving a gap for less-resourced languages such as Italian. This paper introduces Igea, the first decoder-only language model designed explicitly for biomedical text generation in Italian. Built on the Minerva model and continually pretrained on a diverse corpus of Italian medical texts, Igea is available in three model sizes: 350 million, 1 billion, and 3 billion parameters. The models aim to balance computational efficiency and performance, addressing the challenges of managing the peculiarities of medical terminology in Italian. We evaluate Igea using a mix of in-domain biomedical corpora and general-purpose benchmarks, highlighting its efficacy and retention of general knowledge even after the domain-specific training. This paper discusses the model's development and evaluation, providing a foundation for future advancements in Italian biomedical NLP.

## 1. Introduction

The advent of probabilistic language models has revolutionized various domains, with biomedical natural language processing (NLP) standing out due to its significant impact on healthcare provision and medical research. The ability of these models to understand, process, and generate text from vast biomedical corpora has led to improvements in tasks such as entity recognition, relation extraction, and question answering. However, the majority of this progress has been focused on English-language texts, creating a notable disparity for other languages with fewer resources, such as Italian.

In the Italian context, the scarcity of large and diverse training datasets presents a substantial challenge. General language models like Minerva and Maestrale have made strides in Italian NLP, but they lack the specialization required to handle the nuances of biomedical terminology effectively. Addressing this gap is crucial, as the precision and clarity needed in medical communications are paramount for clinical and research applications in such a high-stakes domain.

In this paper we introduce Igea, a biomedical language model (BLM) built from the ground-up on the Italian language, and that is effective in handling Italian native biomedical text while maintaining its efficiency in terms of computational resources. We built upon the foundation model Minerva, which we then continually trained on Italian native biomedical text, while employing proper provisions to avoid disruption of what was learned during pre-training.

The following sections of this paper detail the previous work in the field, the development and training of the Igea model, and its evaluation. This research work aims to demonstrate the potential of Igea to enhance biomedical text processing and generation in Italian, paving the way for future advancements in multilingual biomedical NLP.

---

[1] corresponding author, tommaso.buonocore@unipv.it

## 2. Previous Work

Significant research has been directed towards constructing and fine-tuning language models specifically for biomedical applications. Pioneering works like BioBERT [1] and BioMedLM [2] have tailored the encoder and decoder transformer architecture to absorb and generate biomedical knowledge from expansive literature databases such as PubMed. These models have shown marked improvements in tasks like entity recognition, relation extraction, and question answering within the biomedical domain. Recent advancements in biomedical language models have introduced a variety of sophisticated models such as MedPaLM [3], Meditron [4], MedGemini [5], and GatorTron [6], which further refine the capabilities of NLP systems within the medical domain, focusing on English.

Concerning Italian language processing, efforts have been more sparse compared to English, primarily due to the relative scarcity of large and diverse training datasets [7]. Generative models like Minerva[2], Maestrale[3] or Modello Italia[4] have been instrumental in advancing the Italian NLP landscape, but they are trained for general language understanding and generation. However, the specificity and complexity of medical terminology in Italian pose additional challenges that general language models are ill-equipped to handle without additional targeted training.

Previous works on Italian biomedical alignment include BioBIT [8], a BERT-based model trained solely on machine-translated PubMed abstracts, that is however only 110 M parameters large and is not designed for text generation. To the best of our knowledge, no Biomedical LMs for text generation in less-resourced languages like Italian have been developed or published before.

## 3. Model Development

The Igea model has been developed continually pretraining Minerva, a Transformer model based on the Mistral architecture [9] and trained from scratch on 200B tokens (half Italian and half English). Igea is provided in three distinct sizes to address various computational and application needs: a smaller, Chinchilla-optimal [10], 350 million parameter model, a mid-sized 1 billion parameter model, and a larger 3 billion parameter version. This scaling approach allows for a progressive examination of model performance and utility across different computational resources.

Each model size is trained with the same dataset but configured to explore the trade-offs between resource consumption and model performance, reflecting a common practice in the field to accommodate diverse user scenarios from research to deployment in resource-constrained environments. All the models are hosted, and available to use, on the HuggingFace repository[5].

### 3.1 Biomedical Training Data

The corpus for the continual training of Igea merges multiple sources: medical texts extracted from Italian web sources, a curated collection of medical textbooks, and translated PubMed abstracts, amounting to more than 5 billion words in total, distributed as in Table 1. Web sources include customer-facing websites, online forums and doctor-patient public online conversations. The corpus also includes the subset of Italian Wikipedia pages belonging to the "Medicine" portal[6]. This diverse mixture of datasets constituting the corpus is intended to give the model a comprehensive understanding of both the formal scientific lexicon and the layman ways to communicate medical information in Italian.

---

[2] https://huggingface.co/collections/sapienzanlp/minerva-llms-661e6011828fe67de4fe7961
[3] https://huggingface.co/mii-llm/maestrale-chat-v0.2-alpha-sft
[4] https://huggingface.co/collections/sapienzanlp/modello-italia-igenius-6663482c93abed2f1fa5bc56
[5] https://huggingface.co/bmi-labmedinfo
[6] https://it.wikipedia.org/wiki/Portale:Medicina

Table 1: Words distribution across different training sources. Average, min and max are calculated per document.

| Source type | Number of words | Average (Std) | Min | Max |
| --- | --- | --- | --- | --- |
| PubMed Abstracts | $4377 \times 10^6$ | 254 (96) | 2 | 1720 |
| Web | $659 \times 10^6$ | 742 (881) | 4 | $67 \times 10^3$ |
| Medical Textbooks | $2.3 \times 10^6$ | $772 \times 10^3$ ($567 \times 10^3$) | $362 \times 10^3$ | $1.5 \times 10^6$ |

### 3.2 Training Methodology

Training was conducted over a single epoch. Key hyperparameters following optimization were: a learning rate of 5e-5, leveraging Adam optimizer with modified beta values (0.9, 0.95) to stabilize training convergence. The model employed a cosine learning rate scheduler with a warmup ratio of 0.02. We utilized a distributed multi-GPU setup on a single 8xA100 40GB node to accommodate the extensive computational demands, with gradient accumulation to effectively manage memory and enhance batch processing capabilities.

The model has been deployed using only Hugging Face libraries, utilizing PyTorch's bfloat16 precision to balance computational efficiency with memory usage. Device mapping is automated to optimize resource allocation.

### 4. Evaluation

We provide a baseline indication of Igea's performance reporting accuracy and loss on the held-out evaluation set, shown in Figure 1. At the time of writing, there are no established benchmarks specifically tailored for biomedical text in Italian, which makes it problematic to find reliable metrics and to quantify the performance of the models in the biomedical domain and across different medical subdomains. To fill this gap, we employed neural machine translation to create an Italian version of MedMCQA[7], a multiple-choice question answering dataset designed to address real world medical admission exam questions [11]. 12.5% of the training set and 47% of the evaluation set has been discarded due to translation errors, leading to the final size of 169 thousand Italian question-answer pairs. On this dataset, Igea 3B model scores 31.3% accuracy, as shown in Table 2.

The Igea models have also been tested on three evaluation benchmarks available for Italian but targeting general purpose language understanding, namely MMLU [12], ARC [13] and HELLASWAG [14]. The largest Igea model scores 25.5 on MMLU, 28.7 on ARC, and 49.1 on HELLASWAG, as shown in Table 3. While evaluating biomedical-specific models on generic-purpose benchmarks might sound unfair, this is in fact a useful step to understand if general knowledge acquired during pre-training is forgotten or preserved after in-domain training.

### 5. Conclusion

The Igea family of models represents the first attempt to provide a generative language model designed specifically for biomedical text generation in Italian, making it also the first language model aligned to biomedical knowledge specifically targeting a less-resourced language.

---

[7] https://huggingface.co/datasets/Detsutut/medmcqa-ita

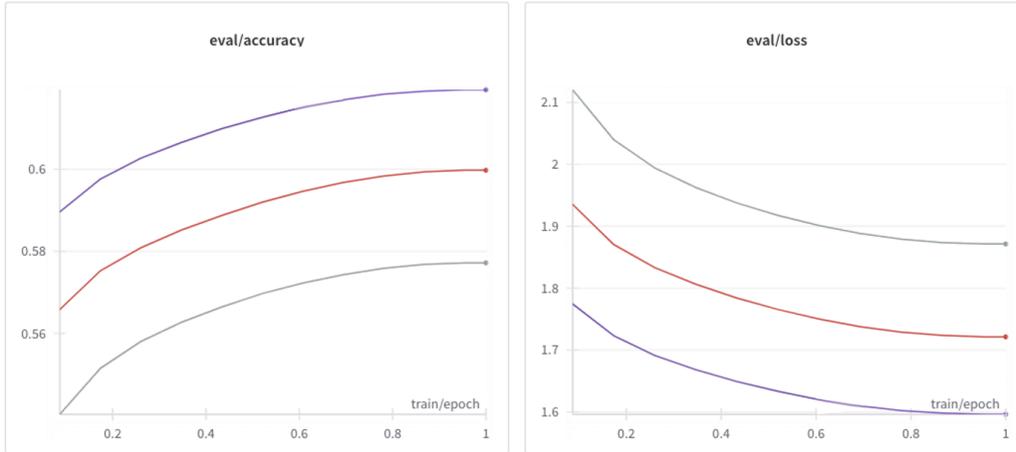

Figure 1: Accuracy and Loss on the held-out evaluation set for Igea 3B (purple), 1B (red) and 350M (grey)

Table 2: Evaluation results in terms of normalized accuracy for the Igea models on the MEDMCQA dataset, translated in Italian. The best checkpoint of Minerva has been included for comparison.

|         |                        | Model                    |           |         |         |
|---------|------------------------|--------------------------|-----------|---------|---------|
|         |                        | Minerva 3B (best base)   | Igea 350M | Igea 1B | **Igea 3B** |
| **Dataset** | MedMCQA-ITA (0-shot) | 29.3                     | 25.0      | 30.7    | **31.3** |

Table 3: Evaluation results in terms of normalized accuracy for the Igea models on the MMLU, ARC, and HELLASWAG datasets, translated in Italian. The best checkpoint of Minerva has been included for comparison.

| Model | Average | MMLU_IT (5-shot) | ARC_IT (0-shot) | HELLASWAG_IT (0-shot) |
|---|---|---|---|---|
| Igea 350M | 26.7 | 25.4 | 24.4 | 30.3 |
| Igea 1B | 29.4 | 25.5 | 27.0 | 35.7 |
| Igea 3B | 34.3 | 25.2 | 28.7 | 49.1 |
| **Minerva 3B** | **36.2** | **26.1** | **30.5** | **51.9** |

Internal validation on loss and accuracy shows that the model benefits from increasing the model size, even when not Chinchilla optimal [10], a requirement that is challenging to meet when constraints exist not only because of the target language but also because of the specific domain of interest. External validation on general-purpose benchmarks shows that common knowledge acquired during the initial pre-training is substantially retained, thus suggesting that no catastrophic forgetting has occurred during the biomedical alignment process.

On the in-domain MedMCQA-ITA dataset, which represents an additional contribution of this paper, Igea always performs better than the corresponding Minerva equivalent, indicating an improvement of the models in terms of biomedical language understanding.

### 5.1. Bias, Risks, and Limitations

The model inherently reflects biases present in its training data, a common issue in pretrained models that can propagate misrepresentations or inaccuracies [15]. Specifically, the model may inadvertently perpetuate stereotypes, biases related to minorities and underrepresented groups, or misinformation through its generated content, a risk that is critical in a medical context. To reinforce responsible use, we distribute the models only through a gating mechanism that ensures that users are informed about the potential ethical implications and the unsuitability of the model for clinical decision-making. Further alignment is thus recommended before usage.

The model may also expose personally identifiable information (PII) leaking from the training data. While the quantity of training data makes it unmanageable to verify the presence of PII in each example manually, we plan to leverage automatic pipelines based on NER to detect and omit sensitive information.

### 5.2. Future Work

Future work requires a continuous refinement of training methodologies, enhancement of dataset quality, and more rigorous evaluation against both language and domain specific benchmarks (when they eventually become available). Next iterations will focus not only on improving the training data mixture, expanding the training corpus with additional diverse medical texts from different sources, but also on enhancing the model evaluation framework by translating existing biomedical benchmarks and adding new datasets to share with the biomedical Italian community.